\documentclass[10pt,twocolumn,letterpaper]{article}

\usepackage[pagenumbers]{cvpr} 

\definecolor{cvprblue}{rgb}{0.21,0.49,0.74}
\usepackage{pifont}
\usepackage{multirow}
\usepackage{algorithm}
\usepackage{algorithmic}
\usepackage{balance}
\usepackage{tabularx}
\usepackage[pagebackref,breaklinks,colorlinks,allcolors=cvprblue]{hyperref}

\def\paperID{9611} 
\def\confName{CVPR}
\def\confYear{2026}

\title{OmniPerson: Unified Identity-Preserving Pedestrian Generation}

\author{
Changxiao Ma$^{1, 2}$, Chao Yuan$^{1}$, Xincheng Shi$^{1}$, Yuzhuo Ma$^{1}$, Yongfei Zhang$^{1, 2, \dagger}$ \\
Longkun Zhou$^{2}$, Yujia Zhang$^{1}$, Shangze Li$^{1}$, Yifan Xu$^{1}$ \\
\textsuperscript{1} Beihang University \quad \textsuperscript{2} Pengcheng Laboratory\\
\textit{Codes: \url{https://github.com/maxiaoxsi/OmniPerson}}
}

\begin{document}
\twocolumn[{%
\renewcommand\twocolumn[1][]{#1}%
\maketitle
\begin{center}
    \centering
    \includegraphics[width=0.95\textwidth]{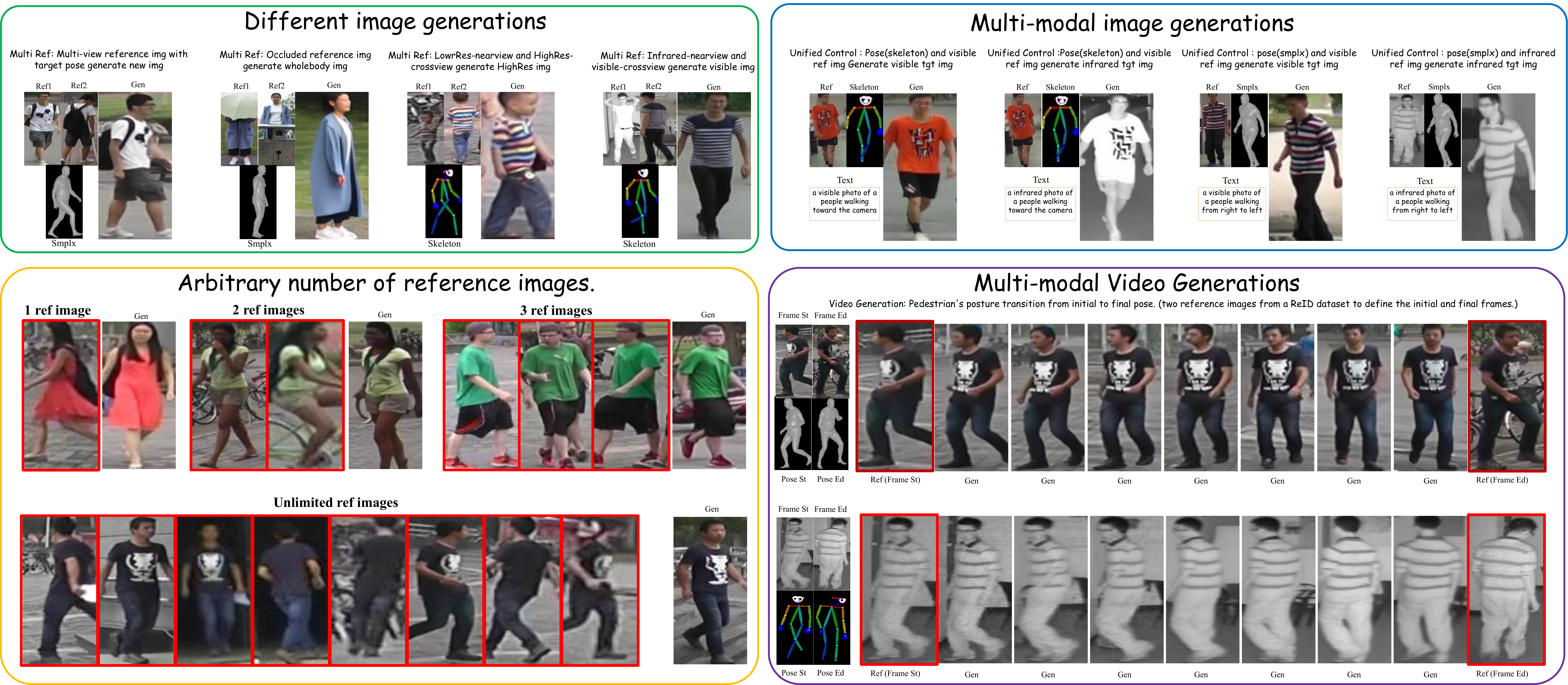} 
    \captionof{figure}{The Muti-tasks generation results of \textbf{OmniPerson}: Controllable Pedestrian Generation from Diverse References.}
\label{fig:qualitative}
\end{center}%
}]
\begin{abstract}
Person re-identification (ReID) suffers from a lack of large-scale high-quality training data due to challenges in data privacy and annotation costs. While previous approaches have explored pedestrian generation for data augmentation, they often fail to ensure identity consistency and suffer from insufficient controllability, thereby limiting their effectiveness in dataset augmentation. To address this, We introduce \textbf{OmniPerson}, the first unified identity-preserving pedestrian generation pipeline for visible/infrared image/video ReID tasks. Our contributions are threefold: 
\ding{192} 
We proposed \textbf{OmniPerson}, a \textbf{unified generation model}, offering holistic and fine-grained control over all key pedestrian attributes. Supporting \textbf{RGB/IR modality image/video generation} with any number of reference images, two kinds of person poses, and text. Also including \textbf{RGB-to-IR transfer} and \textbf{image super-resolution} abilities.
\ding{193} We designed \textbf{Multi-Refer Fuser} for robust \textbf{identity preservation} with any number of reference images as input, making \textbf{OmniPerson} could distill a unified identity from a set of multi-view reference images, ensuring our generated pedestrians achieve high-fidelity pedestrian generation.
\ding{194} We introduce \textbf{PersonSyn}, the first large-scale dataset for \textbf{multi-reference, controllable} pedestrian generation, and present its automated curation pipeline which transforms public, ID-only ReID benchmarks into a richly annotated resource with the \textbf{dense, multi-modal supervision} required for this task. Experimental results demonstrate that OmniPerson achieves SoTA in pedestrian generation, excelling in both visual fidelity and identity consistency. Furthermore, augmenting existing datasets with our generated data consistently improves the performance of ReID models. We will open-source the full codebase, pretrained model, and the PersonSyn dataset.

\end{abstract}


\section{Introduction}

Person re-identification (ReID) aims to retrieve and track target individuals across non-overlapping camera views based on appearance features (e.g. clothing and body shape). In recent years, the field has been dominated by deep learning models\cite{alexey2020image} that have achieved remarkable performance on established benchmarks\cite{li2023clip, He_2021_ICCV}. 

The success of these data-driven methods is, however, critically dependent on the availability of large-scale, diverse, and meticulously annotated datasets. This reliance on massive datasets constitutes a major real-world bottleneck, fraught with challenges including prohibitive annotation costs, stringent data privacy regulations, and the difficulty of capturing the long tail of variations in appearance, pose, and environmental conditions\cite{ristani2016performance}. This data scarcity severely limits the robustness and generalizability of even the SoTA ReID models in real-world deployments\cite{dai2021idm}, motivating the exploration of alternative data sources.

Generative data augmentation has emerged as a powerful strategy to overcome the challenges of data scarcity and limited diversity in ReID. The value of this approach is well-established, as extensive research has demonstrated that augmenting training sets with high-quality synthetic data significantly enhances both the discriminative power \cite{qian2018pose} and generalization capability \cite{Zhang_2021_CVPR} of ReID models. The technology enabling this has progressed rapidly, from early methods using classical image processing \cite{zhong2020random} and virtual engines \cite{Zhang_2021_CVPR}, to the first wave of generative approaches based on GANs \cite{tang2020xinggan}. More recently, the advent of large-scale diffusion models, such as Stable Diffusion \cite{rombach2022high}, has marked a paradigm shift in image synthesis. Capitalizing on their unprecedented realism and control \cite{dhariwal2021diffusion}, the latest works now focus on applying these diffusion-based models specifically for high-fidelity \textbf{person synthesis} \cite{niu2025synthesizing, dai2025diffusion}.

However, despite the strong generative capabilities of these state-of-the-art \textbf{person generation} models, applying them directly to the specialized task of ReID data augmentation reveals two critical limitations.

\begin{figure}[]
\centering
\includegraphics[width=0.9\columnwidth]{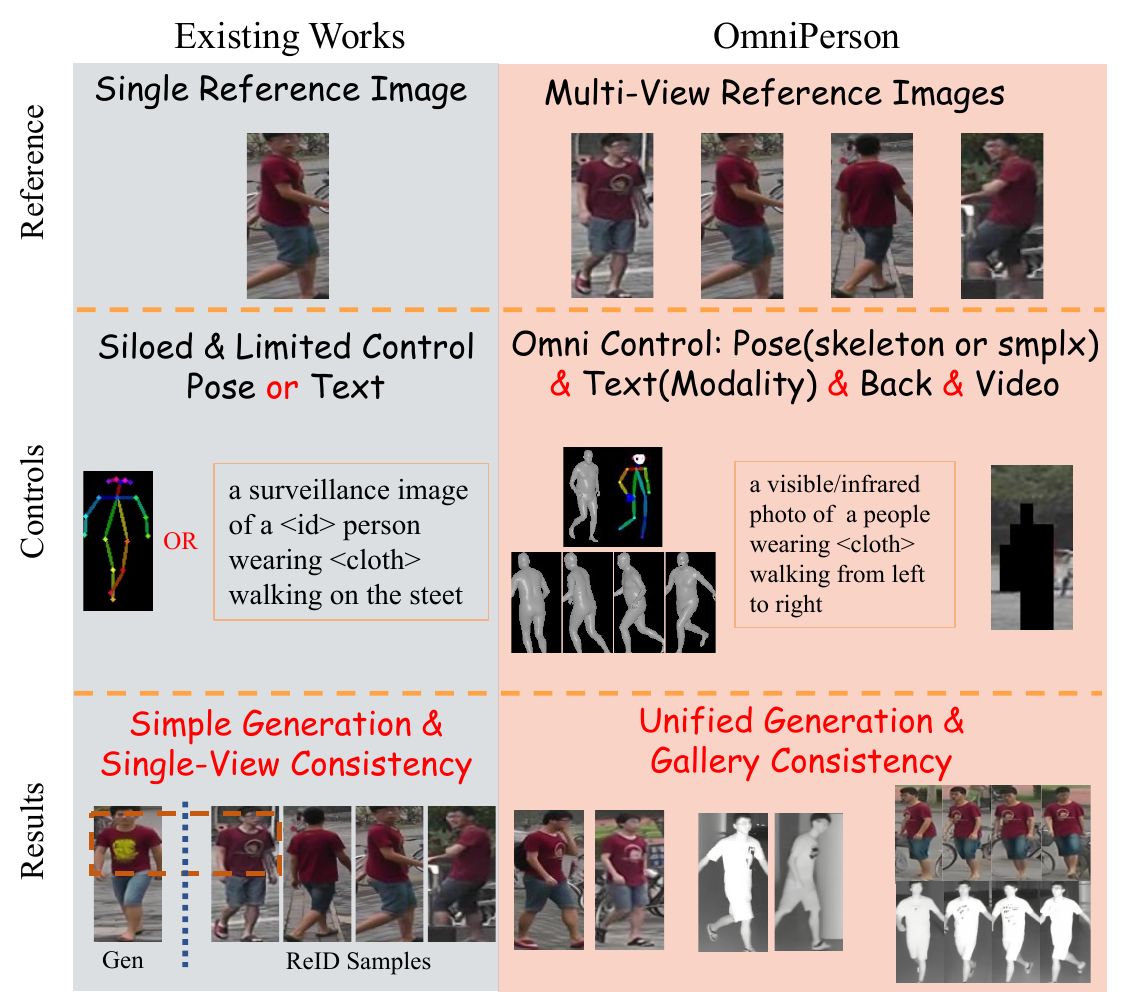} 
\caption{Exiting works vs OmniPerson.}
\label{fig:motivation}
\end{figure}

(1) First, they exhibit a critical failure in \textbf{high-fidelity identity preservation}, especially when generating subjects under drastic variations in pose and viewpoint. The core challenge in ReID data augmentation is creating these novel views, but this is precisely where existing methods, which typically rely on a single reference image for appearance guidance \cite{niu2025synthesizing, dai2025diffusion, Yuan_2025_CVPR}, are fundamentally flawed. As illustrated in Figure~\ref{fig:motivation}, a single view provides an incomplete and often biased representation, failing to capture details visible only from other angles (e.g., the back of a shirt or a logo on a sleeve). Consequently, when tasked with rendering the person from a novel perspective, the model is compelled to hallucinate missing visual information, resulting in significant inconsistencies and "identity drift." This introduces conflicting signals into the ReID training data; instead of learning a robust identity signature, the ReID model is fed ambiguous information that can corrupt its feature space and degrade performance.

(2) Second, they lack a unified and disentangled control mechanism. Existing methods operate in silos: text-driven approaches struggle to faithfully condition on a specific visual identity, often leading to inconsistent results.\cite{saharia2022imagen} Conversely, methods conditioned on a visual reference are often pose-driven, offering little to no explicit control over crucial attributes like background or modality. This fragmented control scheme prevents the targeted synthesis of challenging, long-tail scenarios (e.g., a specific person in a rare pose against a novel background, captured in the infrared spectrum) that are crucial for systematically bolstering the robustness of ReID models.

To overcome these fundamental limitations of identity drift and fragmented control, we introduce \textbf{OmniPerson}, a unified generation pipeline designed specifically for identity-preserving pedestrian data synthesis, and \textbf{PersonSyn}, a detailed reconstructed dataset for pedestrian generation.  Our main contributions are four-fold:
\begin{itemize}
    \item \textbf{A Unified Identity-Preserving Pedestrian Generation model: OmniPerson.} Supporting \textbf{RGB/IR modality image/video generation} with any number of reference images, two kinds of person poses, and text. Also including \textbf{RGB-to-IR transfer} and \textbf{image super-resolution} abilities.
    \item \textbf{Multi-Refer Fuser for OmniPerson.} It achieve the robust \textbf{identity preservation} with any number of reference images as input, making \textbf{OmniPerson} could distill a unified identity from a set of multi-view reference images, ensuring our generated pedestrians achieve high-fidelity pedestrian generation.
    \item \textbf{PersonSyn Dataset.} To facilitate research in multi-reference, controllable pedestrian generation, we reorganized the public pedestrian datasets and obtained \textbf{PersonSyn} dataset with the comprehensive annotations for pedestrian generation, including \textbf{2D/3D poses, disentangled background plates, classified viewpoints, textual attributes,  and reference image sets.} 
    \item Extensive experiments demonstrate that OmniPerson achieves SOTA performance in both generation quality and identity consistency. Furthermore, we validate on downstream tasks and achieve SOTA performance.
\end{itemize}


\section{Related Works}
\label{sec:related_work}

\subsection{Person Re-Identification}

Person re-identification has become a cornerstone task in computer vision, aiming to associate pedestrian identities across non-overlapping camera views and varying time instances. Recent advances in ReID have been largely driven by deep learning, with convolutional neural networks (CNNs) \cite{krizhevsky2012imagenet} and vision transformers (ViTs) \cite{alexey2020image} proving particularly effective in boosting recognition accuracy. State-of-the-art approaches typically integrate feature extraction with metric learning frameworks \cite{liu2025looking,layne2012person, sun2017svdnet, liao2020interpretable, zhang2023protohpe, yuan2025modality,zhang2023pha,yuan2025neighbor}, enabling the learning of discriminative representations that remain robust to challenges such as pose variations, viewpoint changes, and illumination differences.

Standard datasets such as Market-1501 \cite{zheng2015scalable} have been widely adopted to evaluate ReID algorithms under controlled conditions, while SYSU-MM01 \cite{wu2017rgb} focuses on matching identities between visible and infrared images. Although existing methods demonstrate strong benchmark performance on these datasets, the limited number of images and identities, along with homogeneity of collection scenarios, result in ReID models lacking robustness in real-world deployments.

\subsection{Data Augmentation for ReID}
To address data scarcity in person ReID, prior works have employed Generative Adversarial Networks (GANs) \cite{goodfellow2014generative} and virtual simulation engines\cite{qiu2016unrealcv} to synthesize pedestrian images\cite{tang2020xinggan, zhu2019progressive} for dataset augmentation. 
While GAN-based approaches have proven effective for ReID model enhancement through style transfer \cite{zhong2018camstyle, dai2018cross, pang2022cross} and pose-guided generation \cite{qian2018pose, ge2018fd}, virtual engine-based methods demonstrate that large-scale synthetic pedestrian data can significantly improve model generalizability \cite{Zhang_2021_CVPR, wang2022cloning, li2021weperson}. However, the drawback of these methods is the insufficient quality and realism of the generated images, which limits their effectiveness. This fidelity gap has motivated the exploration of more powerful generative frameworks. 

\subsection{Diffusion Model}
The field of image generation has witnessed advancement with diffusion models \cite{ho2020denoising}. Stable Diffusion \cite{rombach2022high} enables high-quality synthesis via latent diffusion, while text conditioning \cite{ramesh2022hierarchical,saharia2022photorealistic} enhances controllability. ControlNet \cite{zhang2023adding} further introduces conditional constraints into diffusion architectures.


Extensive research has demonstrated that diffusion-based methods can generate high-quality pedestrian images \cite{bhunia2023person, Yuan_2025_CVPR}. Subsequent work \cite{niu2025synthesizing} has further validated that Stable Diffusion-based generation can effectively augment ReID datasets. However, despite significant quality improvements over earlier approaches \cite{dhariwal2021diffusion}, maintaining identity-consistent controllable generation remains an open challenge for diffusion-based methods.

Our proposed OmniPerson addresses these limitations through a multi-reference diffusion framework that achieves both identity-consistent generation and simultaneous modality, pose, and background control for ReID data augmentation.
\begin{figure*}
\centering
\includegraphics[width=0.90\textwidth]{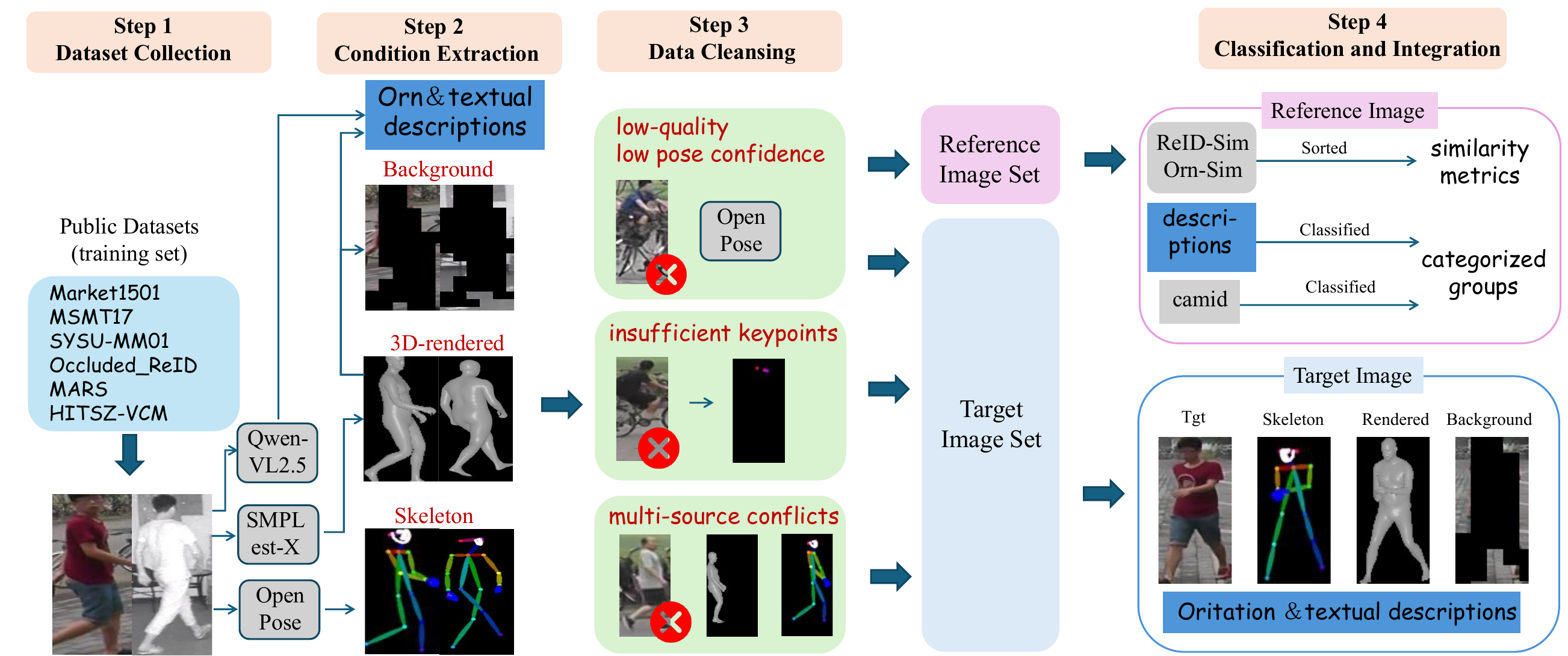}
\caption{The pipeline of PersonSyn dataset generating with all training set of public datasets. }
\label{fig:personsyn_pipeline}
\end{figure*}
\section{The PersonSyn Dataset}
Public ReID datasets lack the rich annotations needed for controllable person generation as they only provide identity supervision. We introduce PersonSyn, a large-scale dataset that transforms these weakly-labeled benchmarks by adding fine-grained annotations for pose, viewpoint, background, visual attributes, and a structured reference set. We now detail our curation pipeline, the automated process designed to generate the comprehensive annotations for the PersonSyn dataset.

\paragraph{Condition Extraction}
The initial Condition Extraction stage derives a rich set of multi-modal annotations from each pedestrian image. We employ OpenPose~\cite{cao2017realtime} to extract 2D skeletal keypoints and concurrently use SMPLest-X~\cite{cai2023smpler} to recover 3D human mesh (SMPL-X~\cite{pavlakos2019expressive}) parameters. These SMPL-X parameters are then leveraged for three key tasks: to render a 3D model, to compute the pedestrian's orientation unit vector, and to enable background extraction. In parallel, semantic attributes are extracted using the Qwen-VL vision-language model\cite{wang2024qwen2}. \textbf{A comprehensive list of these attributes is detailed in Supplementary materials.}
\paragraph{Data Cleaning}
To ensure annotation quality, we apply a strict filtering protocol, removing any samples that meet the following criteria:(1) Poor overall visual quality, (2) Low confidence scores from the pose estimators. (3) An insufficient number of detected keypoints (below a threshold). (4) Significant misalignment between the 2D keypoints and the 3D mesh.
\paragraph{Classification and Integration} 
In this stage, we structure the dataset by first partitioning all intra-identity references into subsets based on their orientation, camera ID, and key attributes (e.g., backpack status). We then pre-compute two metrics for each potential reference-target pair: a CLIP-ReID feature similarity score and a viewpoint score derived from the dot product of their orientation unit vectors. 

Crucially, the goal is not to create one fixed reference set. Instead, this process yields a flexible framework of categorized groups and rich similarity metrics, empowering researchers to design their own sampling logic. For instance, a researcher can generate a person with a backpack, using multi reference images of the same person without one and the guidance of a text prompt "person with a backpack". \textbf{More details about PersonSyn and the specific strategy we designed for training OmniPerson is detailed in the Supplementary Materials}. This entire structured collection of images, annotations, and pre-computed metrics constitutes the final PersonSyn dataset, whitch provides a comprehensive package for each image:
\begin{itemize}
\item \textbf{Conditional Inputs:} Paired 2D skeletons, rendered 3D meshes, and isolated backgrounds.
\item \textbf{Fine-Grained Annotations:} Labels for orientation and key semantic attributes (e.g., clothing, backpack status).
\item \textbf{Flexible Reference Framework:} A pre-processed database of all potential references, partitioned into subsets and scored by visual similarity and viewpoint difference.
\end{itemize}

\begin{figure*}
\centering
\includegraphics[width=0.95\textwidth]{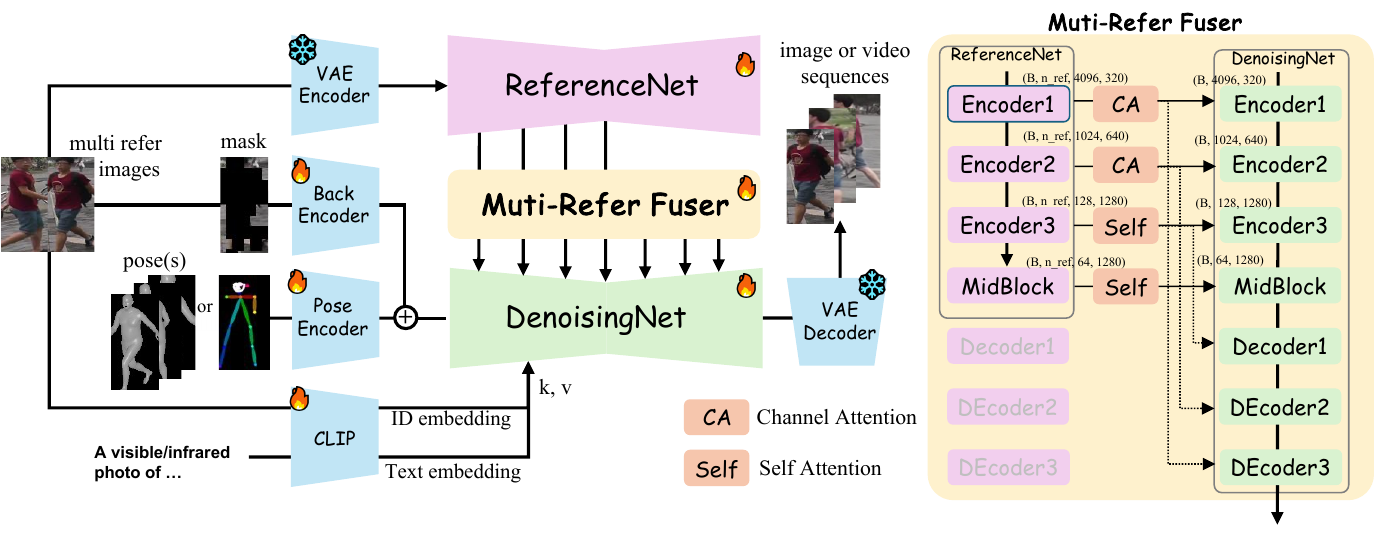}
        \caption{Overview of the \textbf{OmniPerson} framework. Our model leverages the novel Multi-Refer Fuser module to aggregate features from an arbitrary number of reference images. This approach ensures robust gallery-level identity consistency, overcoming the limitations of single-reference methods that are confined to image-level consistency. Furthermore, the framework provides fine-grained, decoupled control over (1) pose (via SMPL-X or skeleton), (2) modality, and (3) background.}
\label{fig:frame}
\end{figure*}

\section{Methods}
This section details the methodology of OmniPerson, our unified generation pipeline. We begin by presenting the overall framework (Sec. \ref{subscec:framework}). Next, we elaborate on our core contribution for identity preservation using multi-view references (Sec. \ref{subsec:identity_preserving}). Finally, we describe the versatile conditioning mechanisms for fine-grained control (Sec. \ref{subsec:omniperson_pipeline}).

\subsection{Unified Generation Framework}
\label{subscec:framework}
Our approach is built upon Latent Diffusion Model (LDM)~\cite{rombach2022high}, which synthesize images $x$ in an efficient latent space. LDM employs a denoising U-Net, $\epsilon_{\theta}$, to iteratively reverse a diffusion process, guided by a condition $c$. This process transforms a random noise vector $z_{t}$ into a clean latent representation $z_{0}$, which is then rendered into a high-resolution image by a VAE decoder$x = D(z_{0})$.

While prior art in pedestrian synthesis often relies on limited guidance (typically a text prompt or a single reference image coupled with pose control). These approaches are insufficient for the structured, fine-grained control required over a pedestrian's holistic identity, pose, background, and modality.

Therefore, our key contribution is to replace this monolithic condition with a unified, multifaceted conditioning framework designed specifically for pedestrian generation.  As illustrated in Figure~\ref{fig:frame}, this framework is guided by a suite of distinct control signals:

\begin{enumerate}
\item A set of $N$ multi-view reference images for identity.
\item A target pose representation(skeleton/smplx).
\item A background scene(optional).
\item A textual descriptor specifying the modality.
\end{enumerate}

The generation pipeline proceeds as follows:
First, the multi-view reference images are processed by our specialized Multi-Ref Fuser Module (see Sec.~\ref{subsec:identity_preserving}) to obtain a robust identity embedding $c_{id}$. Meanwhile, the other conditioning inputs are encoded by their respective encoders to produce embeddings for pose ($c_{pose}$), background ($c_{bg}$), and text ($c_{text}$), as detailed in Sec.~\ref{subsec:omniperson_pipeline}. These embeddings are jointly integrated into the denoising U-Net, guiding the diffusion process to transform random noise $z_t$ into a structured latent code $z_0$ that coherently aligns with all input conditions.

\subsection{Multi-Refer Fusion for Identity Preservation} 
\label{subsec:identity_preserving}

Meaningful data augmentation for ReID requires synthesizing pedestrians across significant variations in pose and viewpoint. This task is fundamentally challenging because a single reference image offers only an incomplete, view-dependent snapshot of a person's appearance, often leading to identity shift during generation. Our approach overcomes this limitation by introducing a mechanism to distill a complete and robust identity representation from the entire set of available reference images for a person. 

The fusion process begins by feeding the set of $N(N \ge 1)$ reference images through a ReferenceNet. To ensure feature space compatibility, this network shares its architecture and weights with the denoising UNet~\cite{Hu_2024_CVPR}. While the ReferenceNet has a full encoder-decoder structure, to balance computational efficiency and sampling quality, followed\cite{zhang2023adding}, we selectively extract features only from its three down-sampling encoder blocks and the middle block. Each U-Net block is composed of a self-attention layer for spatial feature extraction and a cross-attention layer for integrating semantic information. To isolate pure visual information, we disable the cross-attention mechanism and exclusively collect the output features from the self-attention layer in each block. These features serve as the multi-level pedestrian representations for fusion.

Our key insight is that features from different network depths represent different levels of abstraction and thus benefit from distinct fusion strategies. We categorize these features into two types and process them accordingly within our \textbf{Multi-Refer Fuser} module.
\begin{itemize}
    \item \textbf{Low-Level Features}: Extracted from the early stages (down0, down1), these features represent fine-grained visual details such as texture, color, and local patterns.
    \item \textbf{High-Level Features}: Extracted from the deeper stages (down2, mid), these features encode more abstract, semantic information that defines the person's core identity.
\end{itemize}

\textbf{Fusion of Low-Level Features via Channel Attention}. For low-level features, the goal is to select the most salient textural and color information across the reference set. We employ a Channel Attention Fusion mechanism. For a given feature level (e.g., down0), feature maps from all $N$ reference images are concatenated. To capture global channel-wise statistics, we apply both average and max pooling along the spatial dimensions and process the resulting descriptors with a MLP to compute a shared channel-wise attention vector. This vector, passed through a sigmoid function, assigns an importance score to each channel, effectively amplifying informative features (e.g., unique clothing patterns) and suppressing less useful ones. A masked average is then performed across the reference dimension to produce a single, robustly fused low-level feature map.

\textbf{Fusion of High-Level Features via Self-Attention}. For high-level features, it is crucial to understand the contextual relationships between semantic parts across multiple views. To this end, we employ a Self-Attention Fusion mechanism. For a given deep feature level (e.g., mid), the feature tokens from all reference images are concatenated into a single sequence. Standard multi-head self-attention is then applied to this sequence, allowing each feature token to attend to every other token, regardless of its originating image. This enables the model to identify and aggregate the most representative identity-defining features. for instance, a clear frontal view of a face can inform the features of a profile view. A average across the reference dimension yields the final fused semantic feature map.

Finally, the set of fused multi-level features from the RefFuser are integrated into the attention computation of each transformer block: denoising U-Net’s self-attention module augments its key–value set with the fused reference features, enabling the model to perform cross-instance attention. 

Through this mechanism, we extract and fuse multi-level identity features, which are then injected into the denoising UNet to provide a rich and comprehensive representation of the pedestrian's appearance.

\subsection{OmniPerson pipeline}
\label{subsec:omniperson_pipeline}
The versatility of the OmniPerson framework stems from its ability to be guided by a rich, multi-source condition vector, $c$. This condition is a composite of several distinct guidance signals that provide fine-grained control over the generated pedestrian and their environment. In this section, we detail the specific encoders and mechanisms used to process each of these signals: pose, background, textual attributes, and temporal context for video.
\paragraph{Identity Guidance}
While the mechanism detailed in Section~\ref{subsec:identity_preserving} provides a rich and comprehensive representation of the person's visual identity, this alone is insufficient. The denoising UNet needs additional contextual conditioning to properly leverage these features. The goal is to generate a pedestrian that adheres to the target image's specific conditions, like its unique lighting and camera properties.

Our key insight is that standard ReID features, while potentially 'weak' for ReID task, are perfectly suited for this purpose. This is because ReID features are inherently entangled with both camera domain information and high-level pedestrian semantics—precisely the information needed to ground the synthesized image in the desired context.

Therefore, we employ a pretrained CLIP-ReID model to extract a ReID feature from a reference pedestrian (sharing the same ID and camera). This feature, used in conjunction with the text embedding, serves jointly as the key (K) and value (V) for the cross-attention layers within the denoising UNet, thereby guiding the generation of the target pedestrian.

\paragraph{Pose and Background Guidance}
To control the person's structure and the surrounding scene, we employ dedicated spatial encoders for pose and background.

\begin{itemize}
    \item Pose Encoder: This module provides structural guidance. It is designed with the flexibility to process both 2D skeletal keypoints and rendered 3D mesh images. The encoder consists of four convolutional blocks with increasing channel sizes (16, 32, 64, 128), each using strided convolutions to downsample the input. This produces a high-dimensional pose embedding, $E_{pose}$
    \item Background Encoder: To explicitly control the environment, a Background Encoder with an identical architecture processes a desired background image, yielding a spatial embedding, $E_{bg}$, that captures the scene's characteristics.
\end{itemize}

The two spatial embeddings are first combined via element-wise addition to form a single, unified guidance map: $E_{spatial}=E_{pose} + E_{bg}$, and then added directly to the noisy latent $z_t$ before it enters the U-Net.

\paragraph{Text-based Attribute and Modality Control}

OmniPerson also supports semantic control via natural language. We construct a textual descriptor by combining the rich attributes annotated in our PersonSyn dataset (e.g., "a person wearing a blue jacket") with the desired modality (e.g., "a visible photo"). This prompt is then fed into the frozen text encoder of a pretrained CLIP model to generate a powerful semantic embedding, $E_{text}$. This embedding is integrated into the denoising U-Net via the standard cross-attention mechanism, allowing it to guide the overall attributes and modality of the generated image.

\paragraph{Temporal Coherence for Video Synthesis}
To extend OmniPerson from image to video synthesis, followed AnimateAnyone~\cite{Hu_2024_CVPR}, we integrate a temporal coherence module into the U-Net. This module consists of temporal attention layers added to each unet block. When generating the current frame, its intermediate features serve as the query, while the corresponding features from the previously generated frame act as the key and value. To generate videos with continuous pose transformations, particularly across large viewpoint changes, we construct a sequence of intermediate poses to guide the synthesis process, as illustrated in Figure ~\ref{fig:video}. \textbf{The pseudocode for this procedure is provided in the Supplementary Materials}.

\begin{figure}[b]
\centering
\includegraphics[width=0.9\columnwidth]{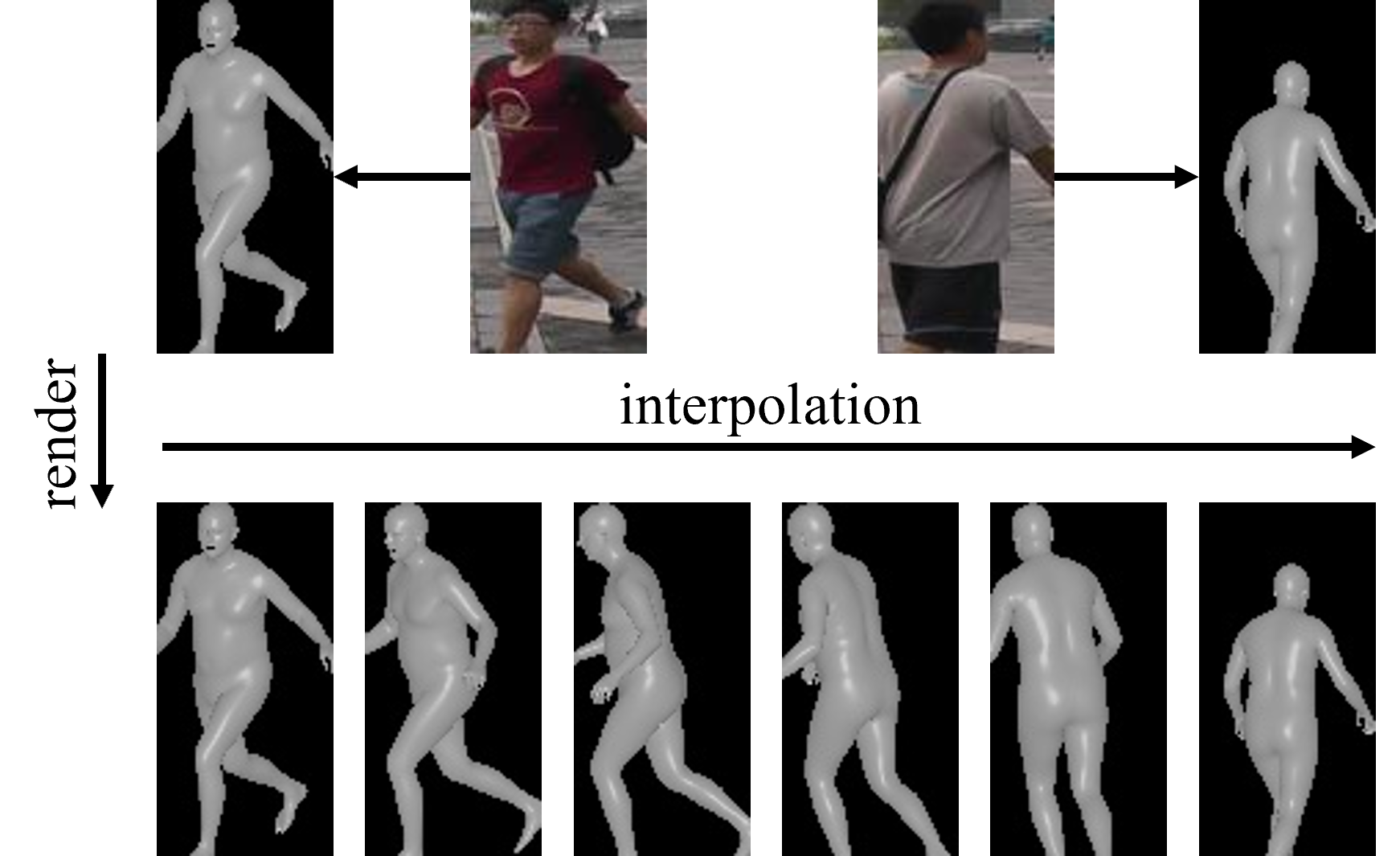} 
\caption{Pose sequence for video generation}
\label{fig:video}
\end{figure}

\paragraph{Training Objective}
The entire OmniPerson framework is optimized end-to-end. The denoising U-Net, $\epsilon_{\theta}$, is trained to predict the noise added to a latent variable, conditioned on the full set of guidance signals. The objective is the standard L2 loss:
\begin{equation}
L = \mathbb{E}{z_{0}, t, \epsilon, c}(||\epsilon - \epsilon_{\theta}(z_{t}, t, c)||_{2}^{2})
\end{equation}
where $c_{id}$, $c_{pose}$, $c_{bg}$, $c_{text}$, $c_{tem}$ represents the condition of identity, pose, background, text, and temporal condition. For video clips, this loss is computed and averaged across all frames. During training, we employ classifier-free guidance by randomly dropping conditions to enhance controllability at inference time.

\section{Experiment}

\subsection{Implementations}
We train OmniPerson on PersonSyn using a single NVIDIA A100 GPU. For each training sample, we curate a reference set by selecting at least one intra-camera image and one image from each orientation set. The selection from each orientation set is probabilistic: candidates are first ranked by ReID similarity, and one is then chosen based on a rank-weighted Bernoulli sampling scheme. Additionally, to increase task difficulty, reference image with the same orientation as the target are probabilistically dropped.   First, we conduct 40K training steps on image datasets with a batch size of 8 and learning rate of $1 \times 10 ^{-5}$. Subsequently, we switch to video training for 40K training steps with a reduced batch size of 2 (processing 8 frames per video) while maintaining the same learning rate. Finally, we fine-tune the model on image datasets for an additional 20K steps with batch size 8, lowering the learning rate to $5 \times 10 ^{-6}$. All input images $I \in R^{w \times h}$(we assume  $\text{h}>\text{w}$ in our formulation, where h and w represent the height and width of the input image) undergo preprocessing where they are first resized to $I \in R^{(\frac{w}{h} \times512) \times 512}$ to preserve aspect ratios, then zero-padded to $I \in R^{512 \times 512}$. \textbf{Detailed data augmentation methods are provided in the supplementary material}.

\subsection{Generated Pedestrian Samples}
Figure~\ref{fig:qualitative} showcases the versatile capabilities of OmniPerson.
The first row demonstrates the model's core strengths in multi-reference fusion and fine-grained controllable generation. This includes complex fusion tasks, such as generating a complete body from occluded inputs or performing cross-modal (Visible-IR) synthesis. Concurrently, it displays decoupled control over pose (dictated by a skeleton or SMPL-X model) and appearance attributes (guided by textual prompts).

The subsequent two rows on the left side of the figure highlight OmniPerson's flexibility in leveraging an arbitrary number of reference images. These examples show successful synthesis using one, two, and a larger set of references, validating the robustness of our fusion mechanism. 

The two rows on the right side showcase our video generation capability. Given start and end frames with a significant pose gap, OmniPerson synthesizes a smooth and temporally coherent video that interpolates the motion between them while preserving identity. \textbf{Additional generation results are provided in the supplementary material, including a qualitative comparison of our method against several baseline approaches.}.

\begin{table}[]
\centering
\small
\renewcommand{\arraystretch}{1}
\caption{Generation Quality Comparison on PersonSyn}
\begin{tabular}{c|cccc}
\hline
Model   & LPIPS$\downarrow$  & SSIM$\uparrow$   & PSNR$\uparrow$    & FVD$\downarrow$    \\ \hline
DSC\cite{siarohin2018deformable}     & 0.303 & 0.305 & 14.308 & -      \\
PATN\cite{zhu2019progressive}    & 0.254 & 0.282 & 14.262 & -      \\
DIAF(\cite{li2019dense})    & 0.306 & 0.305 & 14.201 & -      \\
FD-GAN\cite{dong2020fd}  & 0.612  & 0.210  & 13.981  & -      \\
DIST\cite{ren2020deep}    & 0.282 & 0.281 & 14.337 & -      \\
XingGAN\cite{tang2020xinggan} & 0.306 & 0.304 & 14.446 & -      \\
SPIG\cite{lv2021learning}    & 0.278 & 0.314 & 14.489 & -      \\
DPTN\cite{Zhang_2022_CVPR}    & 0.271 & 0.285 & 14.521 & -      \\ \hline
AnimateAnyone\cite{Hu_2024_CVPR}  & 0.315  & 0.231  & 13.442  & 395.21      \\
Pose2Id\cite{Yuan_2025_CVPR} & 0.309  & 0.236  & 13.900  & 392.1      \\ \hline
Ours    & \textbf{0.202}  & \textbf{0.467}  & \textbf{17.157}  & \textbf{232.34} \\ \hline
\end{tabular}
\label{tab:gen_metric}
\end{table}

\begin{table}[]
\centering
\small
\renewcommand{\arraystretch}{1}
\caption{Generation Quality Comparison on Market-1501}
\begin{tabular}{c|ccc}
\hline
Model   & CLIP-Sim$\uparrow$  & DINO-Sim$\uparrow$   & ReID-Sim$\uparrow$    \\ \hline
DPTN\cite{Zhang_2022_CVPR}    & 0.8636 & 0.6450 & 0.9023   \\ 
AnimateAnyone\cite{Hu_2024_CVPR}  & 0.8661  & 0.6752  & 0.9131    \\
Pose2Id\cite{Yuan_2025_CVPR} & 0.8669  & 0.6825  & 0.9161   \\ \hline
Ours    & \textbf{0.8697}  & \textbf{0.8037}  & \textbf{0.9577}   \\ \hline
\end{tabular}
\label{tab:gen_metric}
\end{table}

\subsection{Generation Comparison with SoTA}
In this section, we compare OmniPerson with state-of-the-art (SoTA) pedestrian generation methods in terms of both generation quality and identity consistency. For each identity in PersonSyn, a target image is randomly selected, with the remaining images serving as the reference pool.

\textbf{For generation quality}, we conduct a comprehensive evaluation using a suite of standard image and video metrics.  Each metric is chosen to assess a distinct aspect of quality:
\begin{itemize}
\item \textbf{LPIPS (Learned Perceptual Image Patch Similarity)} assesses perceptual similarity, closely aligning with human judgments of image quality.
\item \textbf{SSIM and PSNR} evaluate lower-level image fidelity, with SSIM measuring structural similarity and PSNR quantifying pixel-wise reconstruction error.
\item \textbf{FVD (Fréchet Video Distance)} measures the quality and temporal coherence of the generated videos by comparing their feature distributions to those of real videos.
\end{itemize}
To ensure a fair comparison, all methods, including ours, were evaluated without explicit background control. The results clearly indicate that OmniPerson achieves state-of-the-art performance across all key image and video generation quality metrics.

\textbf{For identity consistency}, we measure the feature similarity between the generated images and their ground-truth target samples across three distinct feature spaces: CLIP, DINOv2, and ReID model(pretrained CLIP-ReID). Each metric is chosen to evaluate a different facet of similarity:
\begin{itemize}
\item \textbf{CLIP Similarity} captures the alignment of high-level semantic and stylistic information.
\item \textbf{DINOv2 Similarity} assesses the perceptual and structural correspondence between the images.
\item \textbf{ReID Similarity} directly evaluates the core objective of identity preservation.
\end{itemize}
The experimental results demonstrate that OmniPerson outperforms all existing SoTA methods across every metric, thereby exhibiting the strongest identity consistency.

\begin{table}[]
\small
\centering
\caption{Ablation Study on OmniPerson Components}
\begin{tabular}{cccclll}
\hline
Mul-Ref     & ReID  & Ref-Sel     & LPIPS$\downarrow$  & SSIM$\uparrow$   & PSNR$\uparrow$    \\ \hline
\ding{55}  &   \ding{55} & \ding{55}    & 0.255 & 0.374 & 15.518 \\
\ding{55}  &   \ding{51} & \ding{55}    & 0.234 & 0.420 & 16.661 \\
\ding{51}  &   \ding{55} & \ding{55}    & 0.230 & 0.435 & 16.851 \\ \hline
\ding{51}  &  \ding{51} & \ding{55}     & 0.210 & 0.455 & 16.912 \\
\ding{51}  &   \ding{51} & \ding{51}    & 0.202 & 0.467 & 17.157 \\ \hline
\end{tabular}
\label{lab:ablation}
\end{table}

\begin{figure}[b]
\centering
\includegraphics[width=0.9\columnwidth]{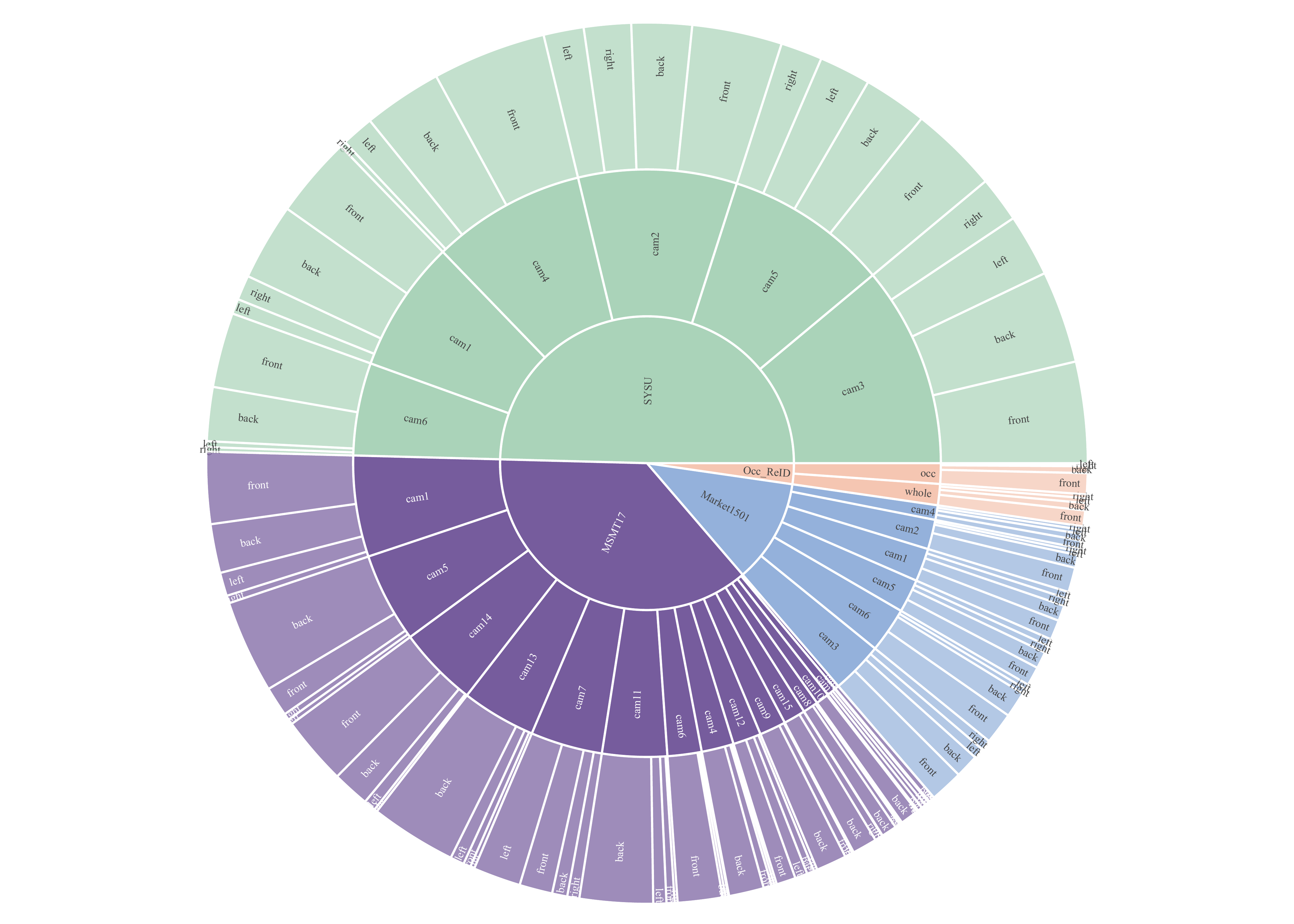} 
\caption{Distribution of cam and view in PersonSyn}
\label{fig:mask}
\end{figure}

\subsection{Ablation Analysis}
The results of our ablation study are presented in Table~\ref{lab:ablation}. We systematically analyze the contribution of three key components: Mul-Ref (our multi-reference fusion mechanism), ReID (the ReID feature guidance), and Ref-Sel (our reference selection strategy for the PersonSyn dataset).

As the results indicate, both the ReID feature guidance and the Multi-Ref fusion module independently improve the quality of the generated images. Furthermore, the experiments demonstrate that when both components are active, employing a well-designed reference selection strategy (Ref-Sel) provides an additional and significant performance boost.

\subsection{ReID Data Augmentation}
To demonstrate the utility of OmniPerson for downstream tasks, we apply it to data augmentation for ReID. We enrich the Market-1501 training set by generating 4 standard-pose images for each identity. As shown in Table~\ref{tab:aug}, training a standard ReID model on data augmented by OmniPerson yields stronger performance improvements compared to using data generated by other methods. It is important to note that our goal is to provide a unified generation tool, not a specialized augmentation method. We therefore used a simple augmentation strategy, which places a higher demand on the generation quality and identity consistency of the synthesized images. Even with this non-optimized approach, OmniPerson delivers stronger augmentation results than competing methods. For fairness, we note that other SoTA methods might also enhance ReID performance, rather than hinder it, if paired with their own tailored augmentation strategies.

\subsection{Statistics and distribution of PersonSyn} 
We present a statistical analysis of the PersonSyn dataset. A subset of these distributions is visualized in Figure~\ref{fig:mask}, \textbf{while a comprehensive breakdown is provided in the Supplementary Materials.}

\begin{table}[]
\small
\centering
\caption{Quantitative performance comparison for pedestrian generation using the Market-1501 dataset.}
\begin{tabular}{l|cc|cc}
\hline
\multicolumn{1}{c|}{\multirow{2}{*}{dataset}} & \multicolumn{2}{c|}{TransReID}    & \multicolumn{2}{c}{CLIP-ReID}     \\ \cline{2-5} 
\multicolumn{1}{c|}{}      & mAP & rank-1 & mAP & rank-1 \\ \hline
Market    & 87.3      & 94.3   & 89.8    & 95.5   \\
Market+DGGAN  & 75.8      & 89.3   & 86.6    & 94.4   \\
Market+DPTN   & 82.6      & 92.2   & 85.3    & 93.3   \\
Market+Pose2Id  & 87.4    & 94.5   & 89.4    & 95.0    \\
Market+ours & \textbf{88.7}  & \textbf{94.7}   & \textbf{90.4}  & \textbf{95.5}  \\ \hline
\end{tabular}
\label{tab:aug}
\end{table}

\section{Conclusion}
In this paper, we propose OmniPerson, the first unified framework for pedestrian generation. OmniPerson integrates multi-reference image to maintain gelleray level identity consistency, while utilizing multi signals to enable visible/infrared pedestrian image/video generation. Furthermore, we introduce and will open-source PersonSyn, a novel multi-modal, multi-reference dataset for pedestrian generation. Experimental results demonstrate that OmniPerson achieves superior generation quality and identity consistency and effectively augments ReID datasets.

\clearpage
\setcounter{page}{1}
\maketitlesupplementary
\appendix

\section{PersonSyn Statistics and Details}
\label{sec:PersonSyn}
We introduce \textbf{PersonSyn}, the pioneering large-scale dataset for \textbf{multi-reference, controllable} pedestrian generation. This section begins with a detailed description of the target image annotations and concludes with the dataset's composition.

\subsection{Annotation}
\begin{figure}[]
\centering
\includegraphics[width=1\columnwidth]{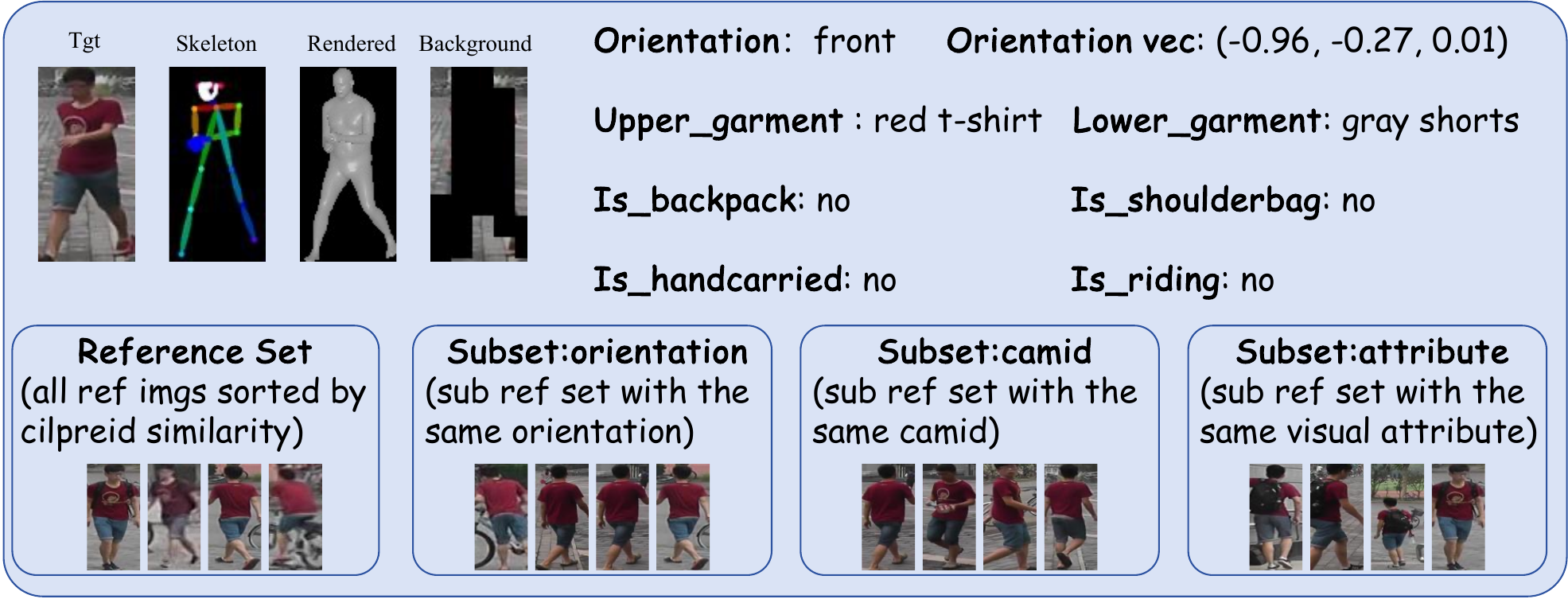} 
\caption{Annotation of PersonSyn}
\label{fig:annotation}
\end{figure}

For each target image, the annotations comprise three main components: conditional inputs, text annotations, and a reference image set. The detailed data annotation structure of PersonSyn is illustrated in Figure \ref{fig:annotation}.
\begin{itemize}
    \item Conditional Inputs: These include the skeleton image, SMPL-X parameters, rendered mesh images, and the background image.
    \item Text Annotations: These cover the orientation description, orientation vector, upper- and lower- garment, as well as attributes indicating whether the pedestrian is carrying a backpack, holding hand-carried items, or cycling.
    \item Reference Image Set: This includes the full collection of reference images (sorted by CLIP-ReID similarity to the target) and their categorizations based on camera identity, orientation, and pedestrian attributes (e.g., carrying a backpack).
\end{itemize}

\subsection{Composition}
\begin{figure*}
\centering
\includegraphics[width=0.95\textwidth]{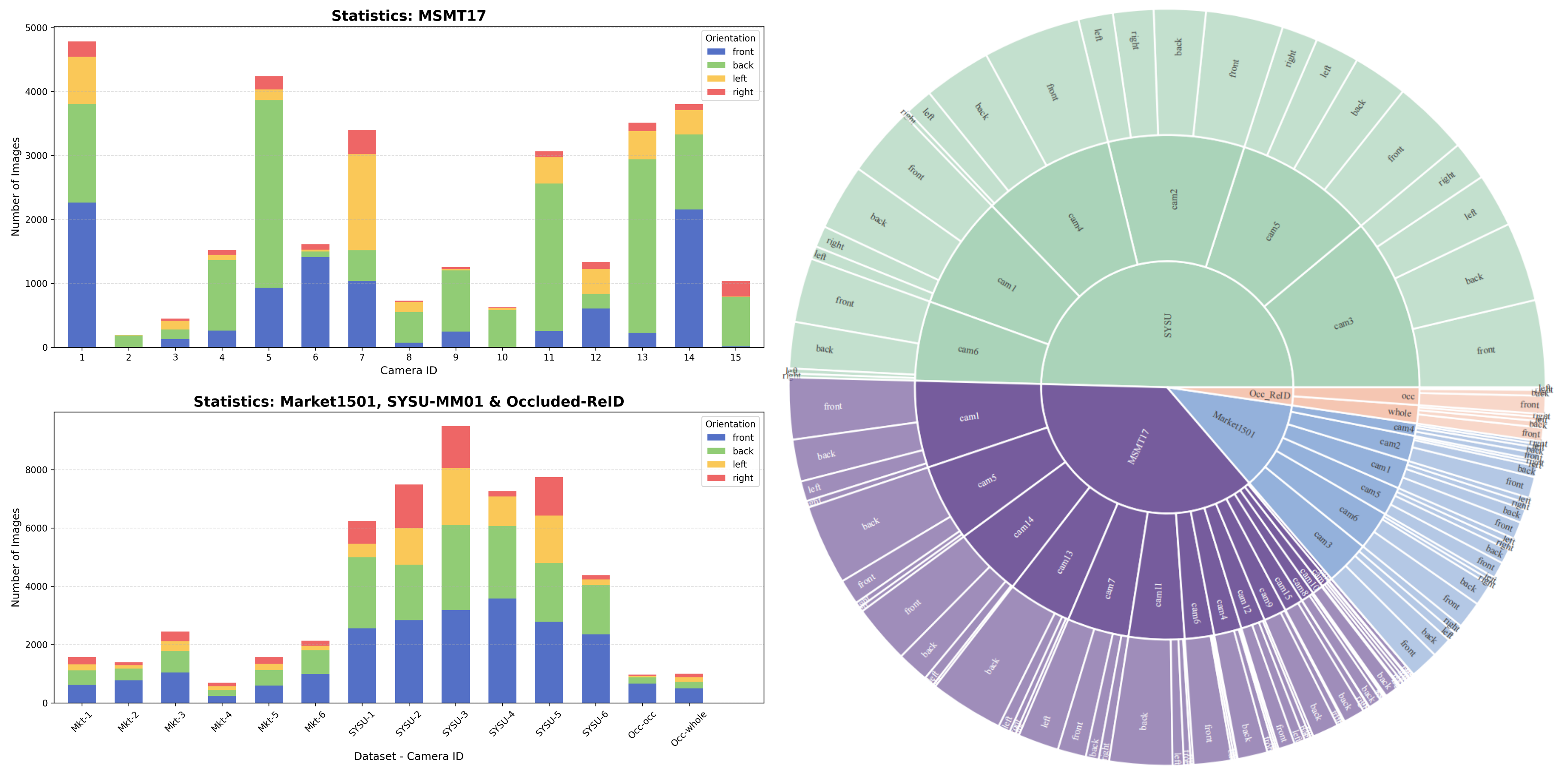}
        \caption{Statistical distribution of camera views and person orientations.}
\label{fig:orientation}
\end{figure*}

PersonSyn is constructed from several public large-scale ReID datasets, including Market1501, MSMT17, SYSU-MM01, Occluded-ReID, MARS, and HITSZ-VCM. These datasets cover a diverse range of tasks: Market1501 and MSMT17 represent traditional image-based ReID; SYSU-MM01 focuses on visible-infrared cross-modality ReID; Occluded-ReID targets occlusion challenges; MARS is designed for video-based ReID; and HITSZ-VCM is tailored for video-based visible-infrared ReID. Specifically, regarding Occluded-ReID, we employ whole-body images as the targets and occluded images as the references. For MARS, the video tracklets are assigned as target images, while Market1501 images are used to provide the reference information. As for HITSZ-VCM, we sample three frames from each video sequence—the start, end, and a random intermediate frame—to serve as the reference images. Detailed statistics on the camera and orientation distributions of the image-based ReID datasets are presented in Table \ref{tab:PersonSyn}, and the visualization results are shown in Figure \ref{fig:orientation}.

\begin{table}[]
\small
\centering
\caption{Statistics of cameras and orientations in image-based ReID datasets.}
\begin{tabular}{p{1.4cm}|c|lllll}
\hline
dataset  & cam   & all & front & back & left & right \\ \hline
\multirow{6}{*}{Market1501}& 1 & 1568 & 625 & 492 & 207 & 244  \\
                           & 2 & 1399 & 775 & 401 & 119 & 104 \\ 
                           & 3 & 2451 & 1043& 743 & 333 & 332 \\ 
                           & 4 & 688  & 244 & 204 & 128 & 112 \\ 
                           & 5 & 1580 & 598 & 528 & 222 & 232 \\ 
                           & 6 & 2130 & 994 & 815 & 155 & 166 \\  \hline
                           
\multirow{15}{*}{MSMT17} & 1 & 4783 & 2263    & 1540 & 740 & 240  \\
                            & 2 & 187  & 1    & 184  & 1   & 1  \\
                            & 3 & 448  & 127  & 151  & 136 & 34  \\
                            & 4 & 1523 & 260  & 1100 & 84  & 79  \\
                            & 5 & 4240 & 931  & 2935 & 166 & 208  \\
                            & 6 & 1623 & 1408 & 90   & 26  & 89  \\
                            & 7 & 3398 & 1040 & 478  & 1500& 380  \\
                            & 8 & 726  & 69   & 481  & 155 & 21  \\
                            & 9 & 1253 & 245  & 960  & 22  & 26  \\
                            & 10 & 625 & 5    & 578  & 29  & 13  \\
                            & 11 & 3064& 254  & 2305 & 412 & 93  \\
                            & 12 & 1335& 607  & 229  & 389 & 110  \\
                            & 13 & 3513& 228  & 2711 & 440 & 134  \\
                            & 14 & 3800& 2156 & 1170 & 381 & 93  \\
                            & 15 & 1037& 13   & 777  & 3   & 244  \\ \hline

\multirow{6}{*}{\shortstack{SYSU-\\MM01}} & 1 & 6240 & 2560 & 2425 & 476  & 779  \\
                           & 2 & 7488 & 2837 & 1901 & 1264 & 1486  \\
                           & 3 & 9502 & 3188 & 2910 & 1968 & 1436  \\
                           & 4 & 7264 & 3581 & 2480 & 1019 & 184  \\
                           & 5 & 7737 & 2785 & 2010 & 1627 & 1315  \\
                           & 6 & 4380 & 2357 & 1698 & 183  & 142  \\ \hline
dataset  & occ   & all & front & back & left & right \\ \hline
\multirow{2}{*}{\shortstack{Occluded-\\ReID}} & occ & 969 & 600 & 214 & 45 & 50  \\
                           & whole & 1000 & 498 & 230 & 153  & 119  \\ \hline
\end{tabular}
\label{tab:PersonSyn}
\end{table}

\section{Qualitative Results and Visualization}
\begin{figure}[]
\centering
\includegraphics[width=0.9\columnwidth]{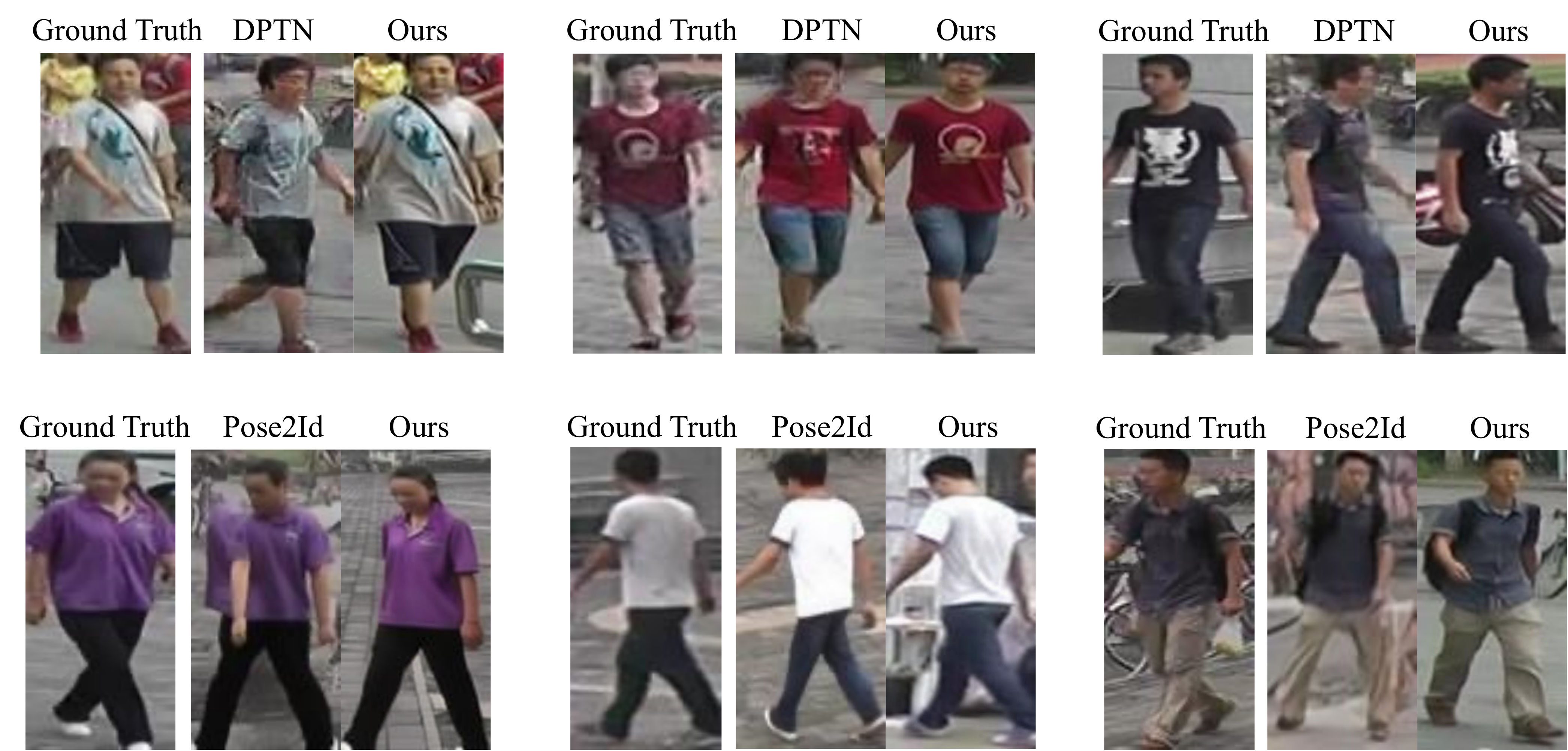} 
\caption{Generation quality comparison: Baseline vs. ours.}
\label{fig:baseline}
\end{figure}

\begin{figure*}
\centering
\includegraphics[width=1\textwidth]{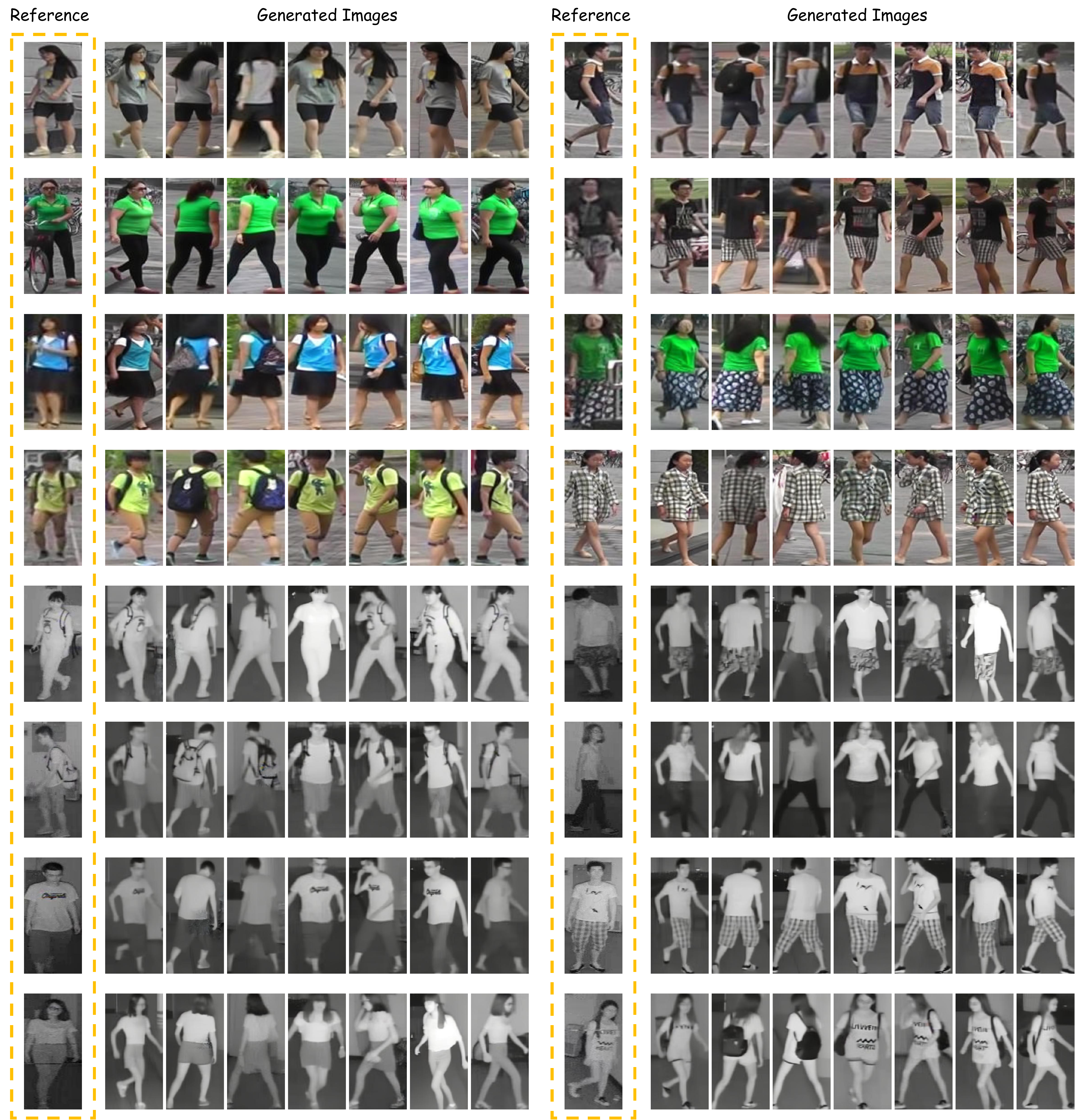}
        \caption{Generation results of multiple identities in predefined poses}
\label{fig:generation_images}
\end{figure*}

\begin{figure*}
\centering
\includegraphics[width=1\textwidth]{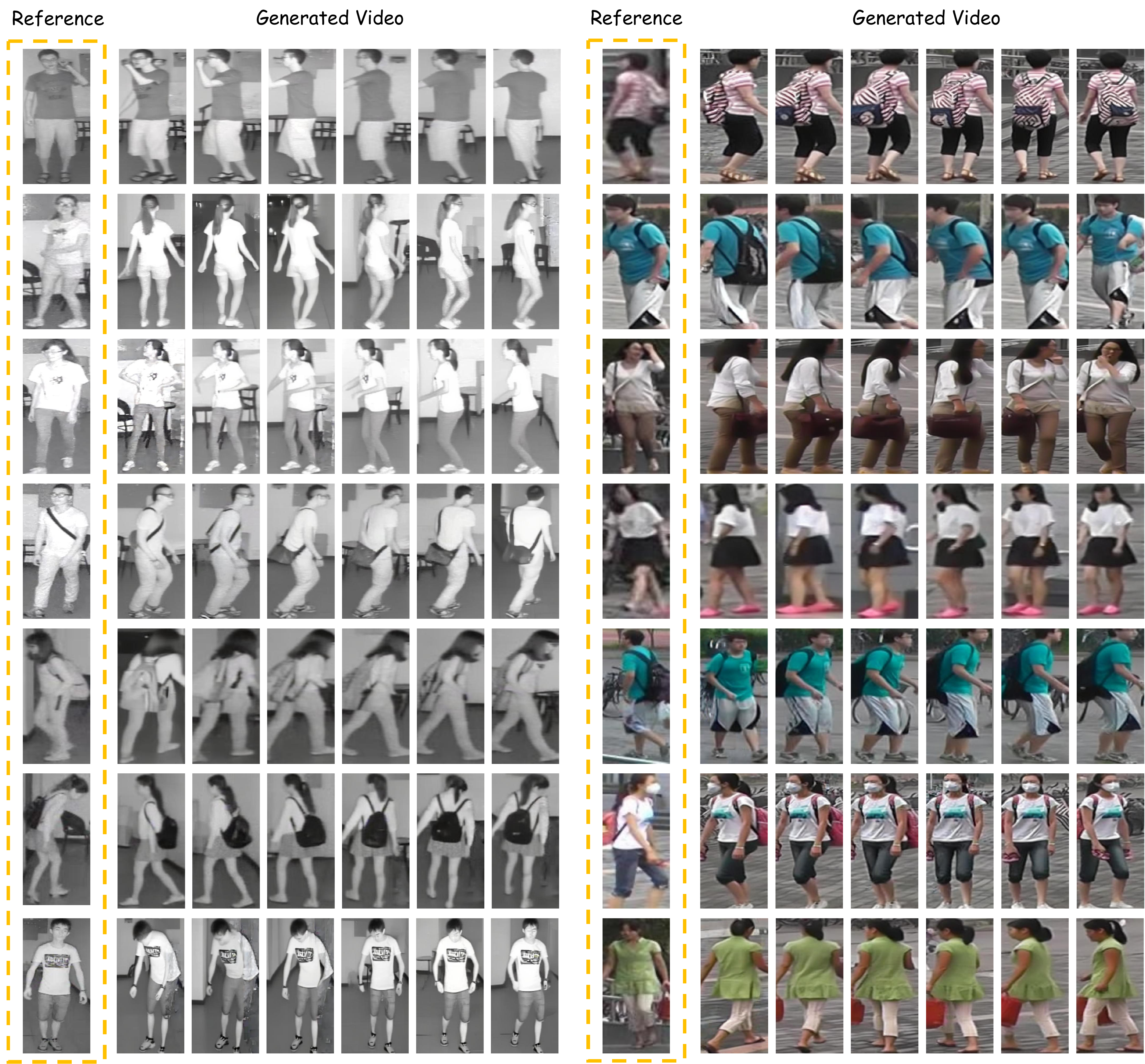}
        \caption{Video generation results of multiple identities}
\label{fig:generation_video}
\end{figure*}

\begin{figure*}
\centering
\includegraphics[width=0.95\textwidth]{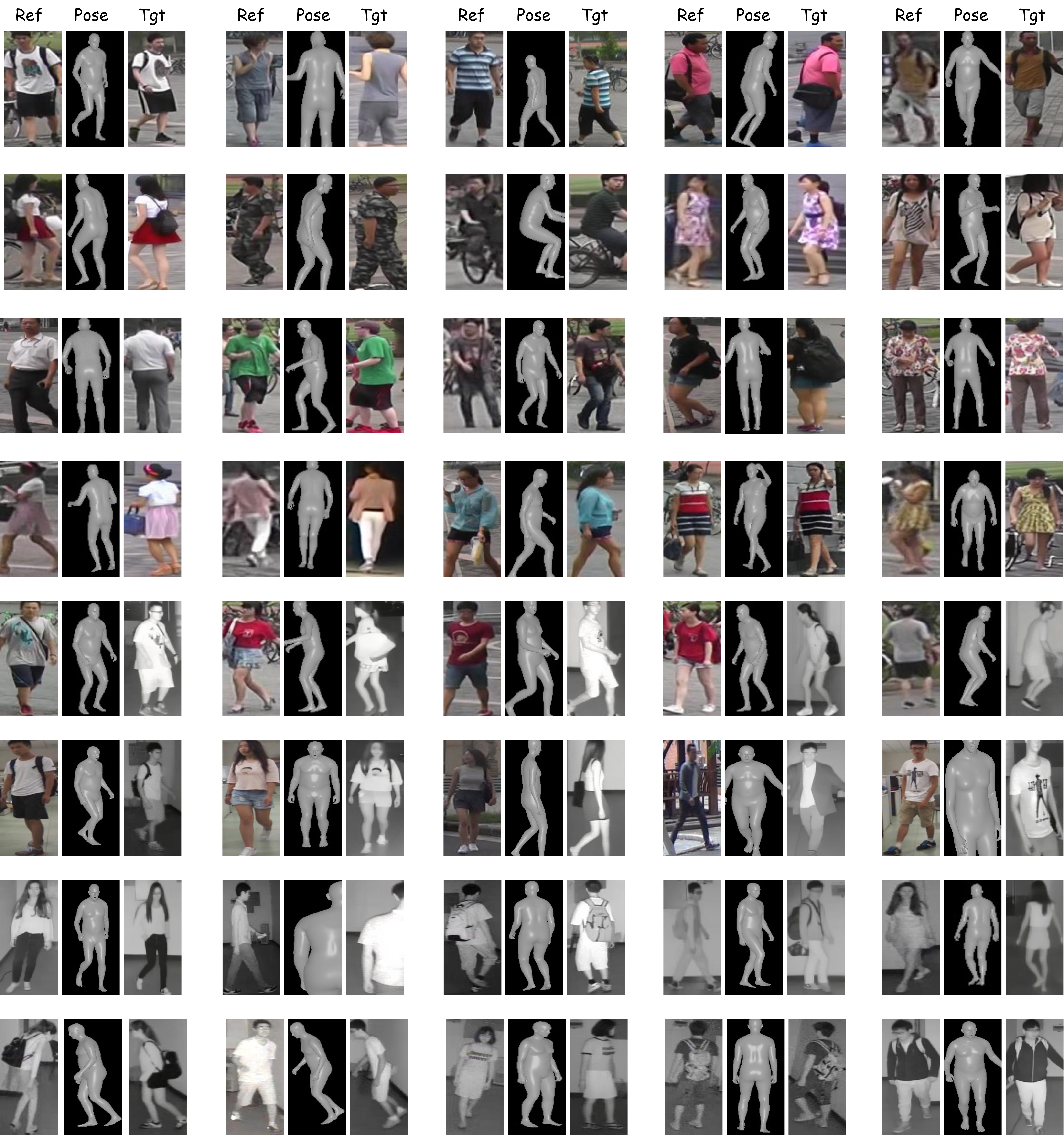}
        \caption{Generation results of random identities in random poses}
\label{fig:generation_random}
\end{figure*}

Figure \ref{fig:baseline} presents a qualitative comparison between OmniPerson and baseline methods. Supplementary generation results are provided in the subsequent figures. Specifically, \textbf{Figure \ref{fig:generation_images} illustrates generation results of multiple identities in predefined poses.} In each group, the leftmost image serves as the real reference (ReID sample), followed by seven synthesized images rendered in fixed standard poses. These visualizations demonstrate that OmniPerson achieves precise control over pedestrian poses and Visible-Infrared modalities while faithfully preserving identity information.
Furthermore, \textbf{Figure \ref{fig:generation_video} showcases video generation results of multiple identities.} Similarly, the first image represents the real reference, followed by six frames sampled from a generated video sequence with continuous pose transitions. This highlights OmniPerson's capability to generate smooth video sequences characterized by natural motion. Notably, while maintaining robust identity consistency, the synthesized videos exhibit significantly larger variations in viewpoint and pose compared to existing real-world Video-ReID datasets. Finally, \textbf{Figure \ref{fig:generation_random} displays generation results of random identities in random poses.} In each group, the first image serves as the real ReID reference, followed by a randomly generated target pose in the middle, and the final synthesized image. These generated results encompass visible image generation, visible-to-infrared generation, and infrared-to-infrared generation, demonstrating the high diversity and coverage of our generative model.

As a multi-reference generation framework, the primary advantage of OmniPerson over the baseline is its ability to maintain gallery consistency via multiple references, rather than merely preserving consistency with a single reference image. Ablation studies on the number of reference images are presented in Section \ref{subsec:ref_ablation}

\section{Algorithmic and Implementation Details}

\subsection{Pedestrian orientation vector calculation}
We extract and compute orientation unit vector from smpl parameters. The rotation matrix is given by the Rodrigues formula:
\begin{equation}
\mathbf{R} = \mathbf{I} + \sin\theta \, [\mathbf{n}]_\times + (1 - \cos\theta) \, [\mathbf{n}]_\times^2
\end{equation}
where $\theta$ is smpl root pose and matrix $[\mathbf{n}]_\times$ defined as:
\begin{equation}
[\mathbf{n}]_\times = 
\left[
\begin{array}{ccc}
0 & -n_{z} & n_{y} \\
n_{z} & 0 & -n_{x} \\
-n_{y} & n_{x} & 0
\end{array}
\right]
\end{equation}
The final pedestrian orientation unit vector is obtained as:
\begin{equation}
\mathbf{v}_{\text{ori}} = 
\left[
\begin{array}{ccc}
r_{11} & r_{12} & r_{13} \\
r_{21} & r_{22} & r_{23} \\
r_{31} & r_{32} & r_{33}
\end{array}
\right]
\left[
\begin{array}{c}
0 \\
0 \\
1
\end{array}
\right]
=
\left[
\begin{array}{c}
r_{13} \\
r_{23} \\
r_{33}
\end{array}
\right]
\end{equation}

\subsection{Image-guided sequence interpolation method}
We generate gradually varying pedestrian parameters through SMPL parameter interpolation and render them to obtain pedestrian video guidance sequences. The interpolation pseudocode is provided in Algorithm \ref{alg:slerp_interpolation}
\begin{algorithm}[tb]
\caption{SMPL Parameters Interpolation}
\label{alg:slerp_interpolation}
\textbf{Input}: SMPL-X parameters $\mathbf{para}_1$, $\mathbf{para}_2$; interpolation coefficient $\alpha \in [0,1]$ \\
\textbf{Output}:paras $\mathbf{para}_{\text{interp}}$
\begin{algorithmic}[1]
\STATE Initialize empty parameter dictionary $\mathbf{para}_{\text{interp}}$

\FOR{each key $k$ in $\mathbf{para}_1$}
    \IF{$k$ is 'global\_orient'}
        \STATE \COMMENT{Handle rotation with quaternion SLERP}
        \STATE Convert $\mathbf{para}_1[k]$, $\mathbf{para}_2[k]$ to rotation vectors $\mathbf{r}_1$, $\mathbf{r}_2$
        \STATE Convert $\mathbf{r}_1$, $\mathbf{r}_2$ to quaternions $\mathbf{q}_1$, $\mathbf{q}_2$
        
        \IF{$\mathbf{q}_1 \cdot \mathbf{q}_2 < 0$}
            \STATE $\mathbf{q}_2 \leftarrow -\mathbf{q}_2$ \COMMENT{Ensure shortest path}
        \ENDIF
        
        \STATE Compute spherical interpolation: 
        $\mathbf{q}_{\text{interp}} = \frac{\sin[(1-\alpha)\theta]}{\sin\theta}\mathbf{q}_1 + \frac{\sin[\alpha\theta]}{\sin\theta}\mathbf{q}_2$
        \STATE where $\theta = \arccos(\mathbf{q}_1 \cdot \mathbf{q}_2)$
        
        \STATE Convert $\mathbf{q}_{\text{interp}}$ back to rotation vector
        \STATE Store result in $\mathbf{para}_{\text{interp}}[k]$
    \ELSE
        \STATE \COMMENT{Linear interpolation for other parameters}
        \STATE $\mathbf{para}_{\text{interp}}[k] \leftarrow (1-\alpha)\mathbf{para}_1[k] + \alpha\mathbf{para}_2[k]$
    \ENDIF
\ENDFOR

\STATE \textbf{return} $\mathbf{para}_{\text{interp}}$
\end{algorithmic}
\vspace{-1mm}
\end{algorithm}

\subsection{Reference Image Selection Strategy.}
During training, we select reference images based on the following criteria:
\begin{itemize}
    \item \textbf{Same-camid Reference}: We explicitly select at least one image from the same camera as the target. Beyond serving as a visual reference, this image is utilized to extract ReID features for guiding generation and to extract the background, which serves as a background prompt.
    \item \textbf{Dynamic Selection}: For the remaining references, we adopt a stage-aware strategy. In the early training stage, we select images with the highest CLIP-ReID similarity to the target from each orientation group. In the later stage, we impose a viewpoint constraint (e.g., requiring a difference of at least $60^{\circ}$ from the target view) to filter the candidate pool. The candidates are then sorted by ReID similarity, and multiple references are selected based on Bernoulli probabilities. 
\end{itemize}

\subsection{Data Augmentation}
For the ReID data used to train the diffusion model, we implement the following augmentation techniques:
\begin{itemize}
    \item \textbf{Random cropping} is applied to both reference and target images before other transformations.
    \item \textbf{Random occlusion} is introduced specifically to the reference images.
    \item \textbf{Background masking} is performed with a high probability (i.e., randomly masking out the background image).
\end{itemize}

\section{Additional Quantitative Evaluations}

\subsection{Visual-Infrared ReID Data Augentation}
As illustrated in Table \ref{tab:aug}, we generated a video sequence with continuously changing actions for each pedestrian under every camera view in the SYSU-MM01 dataset and incorporated them into the training set. We then trained the ReID model and evaluated its performance under the 'all-search' mode. Experimental results demonstrate that the visible-infrared data generated by OmniPerson effectively enhances VI-ReID performance.

\begin{table}[]
\tiny
\centering
\renewcommand{\arraystretch}{1.3} 

\caption{Effectiveness of data augmentation on SYSU-MM01}
\label{tab:aug}

\newcolumntype{Y}{>{\centering\arraybackslash}X}

\begin{tabularx}{\linewidth}{c|c|YYYYY}
\hline
Method                     & dataset   &R-1$\uparrow$ & R-10$\uparrow$ & R-20$\uparrow$ & mAP$\uparrow$ & mInp$\uparrow$ \\ \hline
AGW\cite{ye2021deep}         & SYSU-MM01 & 47.5 & 84.4 & 92.1 & 47.7 & 35.3 \\
HAT\cite{ye2020visible}      & SYSU-MM01 & 55.3 & 92.1 & 97.4 & 53.9 & - \\
LBA\cite{park2021learning}   & SYSU-MM01 & 55.4 & 91.1 & -    & 54.1 & - \\
NFS\cite{chen2021neural}     & SYSU-MM01 & 56.9 & 91.3 & 96.5 & 55.5 & - \\
MSO\cite{gao2021mso}         & SYSU-MM01 & 58.7 & 92.1 & 97.2 & 56.4 & 42.0 \\
CM-NAS\cite{fu2021cm}        & SYSU-MM01 & 62.0 & 92.9 & 97.3 & 60.0 & - \\
MID\cite{huang2022modality}  & SYSU-MM01 & 60.3 & 92.9 & -    & 95.4 & - \\
SPOT\cite{chen2022structure} & SYSU-MM01 & 65.3 & 92.7 & 97.0 & 62.3 & 48.9 \\
FMCNet\cite{zhang2022fmcnet} & SYSU-MM01 & 66.3 & -    & -    & 62.5 & - \\ \hline
\multirow{2}{*}{PMT\cite{lu2023learning}} & SYSU-MM01 & 67.5 & 95.4 & 98.6 & 64.9 & 51.9  \\ 
                           & SYSU+ours & \textbf{68.3} & \textbf{96.0} & \textbf{99.0} & \textbf{66.7} & \textbf{54.1} \\ \hline
\end{tabularx}
\vspace{-3mm}
\end{table}


\subsection{Ablation for Multi Reference Images}
\label{subsec:ref_ablation}
Table \ref{lab:ref_ablation} presents the quantitative metrics for ReID generation across varying maximum reference limits. As shown in the table, a higher number of reference images contributes significantly to better generation quality. We attribute this improvement to the gallery consistency guaranteed by our multi-reference mechanism. By avoiding the bias inherent in aligning with a single reference image, the model achieves higher fidelity. Furthermore, this mechanism allows for the inclusion of historically generated images into the reference set, ensuring consistency across the entire generation sequence.

\begin{table}[]
\small
\centering
\caption{Ablation Study on the Maximum Number of Ref-Img}
\begin{tabular}{c|lll}
\hline
Max-Ref-Img &   LPIPS$\downarrow$  & SSIM$\uparrow$   & PSNR$\uparrow$    \\ \hline
1    & 0.270 & 0.284 & 13.833 \\
2    & 0.235 & 0.359 & 15.625 \\
3    & 0.216 & 0.437 & 16.922 \\
4    & 0.210 & 0.455 & 16.912 \\ \hline
\end{tabular}
\label{lab:ref_ablation}
\end{table}

\clearpage

\clearpage

{
    \small
    \bibliographystyle{ieeenat_fullname}
    \bibliography{main}
}

\end{document}